\documentclass[sigconf]{acmart}

\usepackage{enumitem}
\usepackage{CJKutf8}
\usepackage{amsmath}
\usepackage{bbm}
\usepackage{multirow}
\usepackage{bbding}
\usepackage{xcolor}
\usepackage{colortbl}
\AtBeginDocument{%
	\providecommand\BibTeX{{%
			\normalfont B\kern-0.5em{\scshape i\kern-0.25em b}\kern-0.8em\TeX}}}

\setcopyright{acmcopyright}
\copyrightyear{2018}
\acmYear{2018}
\acmDOI{10.1145/1122445.1122456}

\acmConference[Woodstock '18]{Woodstock '18: ACM Symposium on Neural
	Gaze Detection}{June 03--05, 2018}{Woodstock, NY}
\acmBooktitle{Woodstock '18: ACM Symposium on Neural Gaze Detection,
	June 03--05, 2018, Woodstock, NY}
\acmPrice{15.00}
\acmISBN{978-1-4503-XXXX-X/18/06}

\begin{document}
	
	\title{ComQA:Compositional Question Answering via \\ Hierarchical Graph Neural Networks}
	
	\author{Bingning Wang$ ^{1} $, Ting Yao$ ^1$, Weipeng Chen$ ^1$, Jingfang Xu$ ^1$ and Xiaochuan Wang$ ^{1,2}$}
	\affiliation{%
		\institution{1. Sogou Inc. $ \quad $   2. Tsinghua University.}
		\city{Beijing}
		\country{China}
	}	
	\email{wangbingning,yaoting@sogou-inc.com}
	
	

	\begin{abstract}
		With the development of deep learning techniques and large scale datasets, the question answering (QA) systems have been quickly improved, providing more accurate and satisfying answers. However, current QA systems either focus on the sentence-level answer, i.e., answer selection, or phrase-level answer, i.e., machine reading comprehension. How to produce compositional answers has not been throughout investigated. In compositional question answering, the systems should assemble several supporting evidence from the document to generate the final answer, which is more difficult than sentence-level or phrase-level QA. In this paper, we present a large-scale compositional question answering dataset containing more than 120k human-labeled questions. The answer in this dataset is composed of discontiguous sentences in the corresponding document. To tackle the ComQA problem, we proposed a hierarchical graph neural networks, which represent the document from the low-level word to the high-level sentence. We also devise a question selection and node selection task for pre-training. Our proposed model achieves a significant improvement over previous machine reading comprehension methods and pre-training methods. Codes, dataset can be found at \url{https://github.com/benywon/ComQA}.
	\end{abstract}


	\maketitle
	
	\section{Introduction}\label{intro}
	Question answering (QA) is a long-term research interest in the NLP community \cite{phillipsartificial,greenwood2005open}. Based on the knowledge resource, QA could be categorized into knowledge-based QA (KBQA), community QA (CQA) or textual QA. In KBQA, the answer is produced from a knowledge base. In CQA, the answer is derived from the user-generated content such as FAQ. In textual QA \cite{harabagiu2003open}, the knowledge resource is unstructured text such as Wikipedia pages or news articles. Textual QA extracts a sentence or answer snippet from a supporting document that responds directly to a query, which is the main form of the current QA system \cite{yogish2017survey}. 
	
	\begin{figure}
		\centering
		\includegraphics[width=1.0\linewidth]{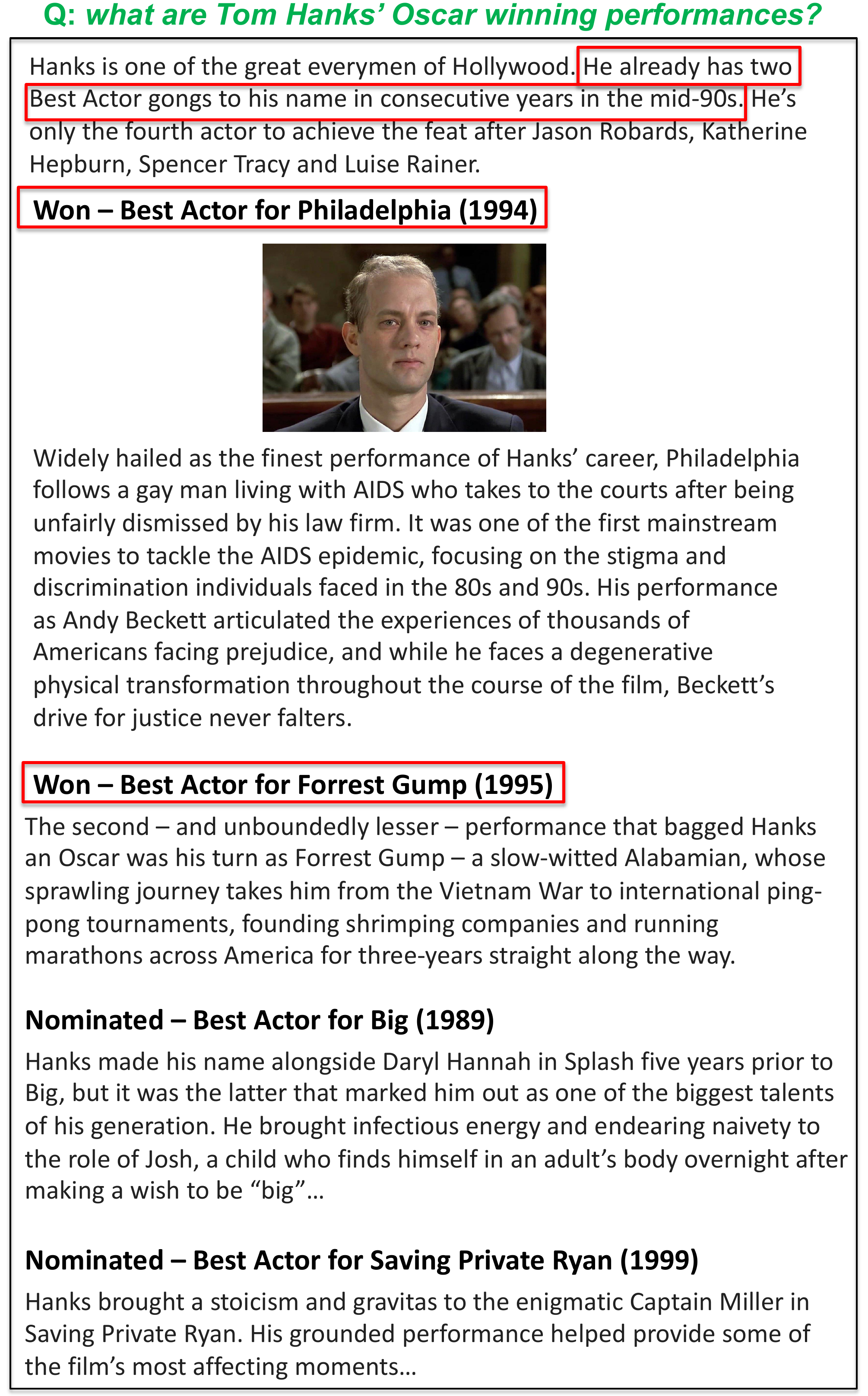}
		\caption{An example of a compositional QA system where the answer lies in discontinuous parts of a document. The query is green and the answers are marked by red frame. }\label{fig_example}
	\end{figure}
	
	Traditional textual QA is mostly based on symbolic feature engineering \cite{cui2005question,verberne2009learning}. Recently, with the development of deep learning methods and large scale datasets, especially the techniques of machine reading comprehension, the textual based QA systems have been greatly improved \cite{Seo2016BidirectionalAF,nguyen2016ms,rajpurkar2016squad,dunn2017searchqa,Dhingra2017QuasarDF,kwiatkowski2019natural}. Most of the recent advances in textual based QA are sentence-level or phrase level. For example, most of the answer selection models can be regarded as two-sentence matching methods, which select the most relevant sentence (or paragraphs) for a query. On the other hand, most machine reading comprehension methods resort to span extraction to extract the consecutive answer span in a passage given the query, where the answer is usually a very short phrase \cite{baradaran2020survey,zeng2020survey}.
	
	However, for the current textual QA applications, some of the answers are compositional, i.e., they are made up of discontinuous sentences in a document. Take the case in Figure \ref{fig_example} for example\footnote{https://www.standard.co.uk/go/london/film/tom-hanks-oscars-wins-nominations-forrest-gump-a4334001.html}, given the query asking about Tom Hanks' Oscars winning movies, the answer is made up of discontinuous parts in the page. Concretely, the answer is composed of three sentences-two from the subtitle and one from a sentence in the first paragraph. Current QA systems, focus on either sentence-level or phrase-level, are unable to tackle this problem.
	
	In this paper, we are focusing on \underline{Com}positional \underline{Q}uestion \underline{A}nsw-\\ering (ComQA), where the answer is from the discontiguous but relevant sentences in a document. For the currently available QA datasets, none of them are focus on compositional answers, so we construct a large scale Chinese compositional QA dataset through crowdsourcing. First of all, we select the web pages whose title is a valid question. Then we conduct the page analysis to extract the main content of the page as the document. Next, we transform the original document HTML to a list of sentence-like items that each item can serve as a candidate component of the final answer. Finally, we feed the question and the transformed page to the crowd workers to select the answers. 
	
	Compared with previous textual based QA, the ComQA has three characteristics: 1) Rather than restricting the answer to be a span or a sentence of the text, ComQA loosens the constraints so that the answer may lie in discontiguous components in a document. 2) The basic granularity of ComQA is a sentence. However, every element in the dom-tree of the page's HTML, e.g., a table or an image, could be a valid component to form the final answer. 3) Since the answer in ComQA is composed of different granularity, it introduces the specific inductive bias of the document structure to the model, requiring a hierarchical understanding of the text from the low-level words to high-level paragraphs.
	
%	We conduct several baseline experiments on the proposed ComQA dataset, including the widely used machine reading comprehension model BiDAF \cite{Seo2016BidirectionalAF} and recently proposed pre-training model BERT \cite{devlin2018bert}. The results of those models are inferior compared with human performance. Therefore, 

	To tackle the problem of compositional answers, we proposed a novel hierarchical graph neural networks on the ComQA. Concretely, we first adopt a BERT-like sequential encoder to obtain the basic token representations. Then we represent the question and the elements of the document in a graph. The graph is constructed from dom-tree of the page with different levels of connection. Finally, we represent each node in the graph with the attention-based hierarchical graph neural networks. The final prediction is based on the node representations in the graph.
	
	Focusing on document understanding, we also devise two novel pre-training tasks for our proposed hierarchical graph neural networks: the first one is the question selection task to determine the relationship between the question and the document. The second is the node selection task measuring the contextual dependency of a node in the document. We conduct comprehensive experiments on the ComQA dataset. The proposed attention-based hierarchical graph neural networks achieve significant improvement compared with the state-of-the-art QA baselines, including the widely used machine reading comprehension model BiDAF \cite{Seo2016BidirectionalAF} and recently proposed pre-training model BERT \cite{devlin2018bert}. The ablation study shows the advantage by incorporating the graph structure and pre-training. In summary, our contributions are:
	\begin{enumerate}[leftmargin=*]
		\item This paper investigates the compositional question answering, where the answers are composed of discontiguous supporting sentences in the document.
		\item We present a new human-labeled compositional question answering dataset containing more than 120k QA pairs. To our best knowledge, this is the largest one in this domain.
		\item We propose a hierarchical graph neural networks on the compositional QA, and two novel tasks are devised to pre-train the model. The proposed model achieves significant improvement over previous methods such as BiDAF \cite{Seo2016BidirectionalAF} or BERT \cite{devlin2018bert}.
	\end{enumerate}
	\section{Related Work}
	\textbf{Question Answering}. QA has been a long-term focus in artificial intelligence that can date back to the 1960s, where \citet{green1961baseball} proposed a simple system to answer questions about baseball games. Since then, many works have been done to use diverse data resources to answer any type of question. Compared to other types of QA such as knowledge base QA or community QA, in textual QA the knowledge resource is the plain text such as document or paragraph, which is currently regarded as the machine reading comprehension (MRC). Since the MCTest \cite{richardson2013mctest} was proposed, many researchers have been focused on MRC. \citet{hermann2015teaching} proposed a cloze-style dataset CNN/Daily Mail where the questions and answers are automatically generated from the news article. Recently, some human create large-scale authentic datasets have been released, such as SQuAD \cite{rajpurkar2016squad}, NewsQA \cite{trischler2016newsqa}, etc. However, most answers in these datasets are based on the contiguous segments in the text, and most MRC methods are based on span extraction, which can hardly deal with the compositional answers in QA.
	
	\noindent
	\textbf{Compositional QA}. Some works have also been proposed to solve the compositional answers in QA. In MARCO \cite{nguyen2016ms} dataset, some answers could not be directly derived from the passage, \citet{yang2020multi} and \citet{zhang2020composing} proposed to generate the answer based on the multi-span extraction and scoring scheme. However, their works are heavily based on the text's syntactic features to extract a continuous span. Besides, the answers are not really compositional in MARCO, given that most non-extractive answers are continuous span in the passage combined with a span in the question. A specific dataset called  \cite{dua2019drop} has been proposed containing the multi-span answers. Nevertheless, the compositional answers only account for 6\% of the datasets, and each answer span is short phrases. MultiRC \cite{khashabi2018looking} is a multiple-choice dataset requiring reading comprehension from multiple sentences. Nevertheless, the dataset is too small to train an applicable model, and the multiple evidence sentences are not corresponding to the final answer. Furthermore, in MultiRC the questions are synthetic by humans, while our proposed ComQA consists of real-world questions. 
	Some multi-hop QA datasets have recently been released, such as WikiQA \cite{welbl2018constructing} and HotpotQA \cite{yang2018hotpotqa}, but they are mainly focused on the multi-document reasoning and the answer is continuous span in one document. 
	
	Different from their works, in this paper, we are only focusing on compositional QA, where the answers are derived from multiple but discontiguous segments in the documents. And the proposed ComQA is large-scale, containing more than 120k questions.
	
	\noindent
	\textbf{Graph Neural Networks}. Graph neural network (GNN) is a kind of deep learning method that operates in the graph. It has gained wide attention due to its convincing performance and high interpretability. GNN incorporate the explicit relational inductive bias to the model \cite{battaglia2018relational} to perform structure reasoning. \citet{kipf2016semi} proposed graph convolutional network to conduct semi-supervised learning on the club network. To focus on the important part of the graph, \citet{velivckovic2018graph} proposed the graph attention networks. Many other NLP tasks have also employed the GNN to model the text, such as relation extraction \cite{zhang2018graph}, text classification \cite{yao2019graph}, dependency parsing \cite{ji2019graph}, or question answering \cite{tian2020capturing} etc.
	
	\section{Compositional QA}
	In this section, we will first introduce the ComQA in detail. Next, we will describe the data construction process.
	
	Compositional QA is similar to the machine reading comprehension, where the system produces an answer given the question and a corresponding document. However, in ComQA, the answer is located in different parts of the document, just like the instance illustrated in Figure \ref{fig_example}. In this paper, we classify the answer components into two types:
	
	\noindent
	\textbf{Sentence}: sentence is the basic component in ComQA. The sentence could be a subtitle on the page; for example, the sentence in Figure \ref{fig_example} with a bold and large font. Or intra-paragraph sentences. We use the simple heuristic rules to divide the paragraph into several sentences, and each sentence could be a valid candidate answer component.
	
	\noindent
	\textbf{Special HTML Element}: For many questions in ComQA, the answer may lie in a specific element on the page other than the raw paragraphs. For example, for the question \textit{`How to turn on PS4 motion controller?'} the answer may consist of several images describing the operation process. Another example is when the document is a wiki page of `\textit{List of highest mountains on Earth}', it contains a table of the ranking of the world highest mountains, when asked about the `\textit{third highest mountain in the world}', we should extract the third item in the table as the answer. In this paper, we kept several special HTML tags that potentially compose the answer. They are listed in Table \ref{tab.tags}.

	\renewcommand\arraystretch{1.05} 
	\begin{table}[!ht]
		\begin{tabular}{c|l}
			\multicolumn{1}{c}{Tag} & \multicolumn{1}{c}{Definition}      \\ \hline \hline
			br                      & produces a line break in text       \\ \hline
			p                       & paragraph marker                    \\ \hline
			img                     & image content                       \\ \hline
			table                   & table content                       \\ \hline
			strong                  & define important text in a document \\ \hline
			tbody                   & group the body content              \\ \hline 
			h1-h6                   & define headings \\ \hline \hline          
		\end{tabular}
		\caption{The HTML tags we kept as the special nodes in the document graph.}\label{tab.tags}
	\end{table}

	The answer in ComQA is made up of the above two types of components, which we refer to as \textit{node} in the rest of the paper. Other types of answers, such as phrases and words in the paragraph or video elements in the HTML tree, are also valid for ComQA. However, adding them would increase the tasks' difficulty, so we leave them for future works. In ComQA the system should assemble the discontiguous nodes as the answer.
	
	\subsection{Dataset Collection}
	
	We collect the ComQA dataset based on the Chinese search engine Sogou Search\footnote{\url{https://www.sogou.com/}}. The data construction consists of four stages: First, we obtain the question-document through a search engine. Then we process the documents into nodes. Next, crowdsourcing workers were asked to select the corresponding nodes in the document to answer the question. Finally, we conduct quality inspection on the labeled dataset to filter out invalid samples.
	
	\subsubsection{\textbf{Question-Document Collection}}
	We first obtain the question from the web. We select the pages for which the titles are valid questions and treat the title as the question. We use rules to determine whether a title is a valid question, including: 1) whether the question contains the interrogative pronoun such as \begin{CJK}{UTF8}{gbsn}`谁是'\end{CJK}(`\textit{who}'), \begin{CJK}{UTF8}{gbsn}`如何'\end{CJK}(\textit{`how to'}), etc. 2) the title containing the informative words such as \begin{CJK}{UTF8}{gbsn}`的过程'\end{CJK}(`\textit{the process of}'), \begin{CJK}{UTF8}{gbsn}`的方法'\end{CJK}(`\textit{the method to}'), etc. The question words and informative words are listed in the Appendix. Although the two rules are simple, we found it is really effective that more than 90\% of the selected titles are valid questions.
	
	Next, we also do some filtering process on the selected documents for which the title is a question. We limit the page source to be a list of sites with high quality. Then we filter out the pages containing no text content or just a pure advertising page. We also remove the pages that are either too long or too short. 
	
	After the above two processes, we obtain a lot of high-quality question-document pairs. However, since the web content tends to be duplicated, some of the pages may have highly lexical or semantic overlap. To remove the redundant pages and increase the final dataset's diversity, we represent the page document based on the bag-of-words. We use Ward \cite{ward1963hierarchical} clustering algorithm to cluster those documents into 350k clusters. We select the centroid in each cluster as the final document for crowdsourcing.

	\begin{figure}
		\centering
		\includegraphics[width=0.85\linewidth]{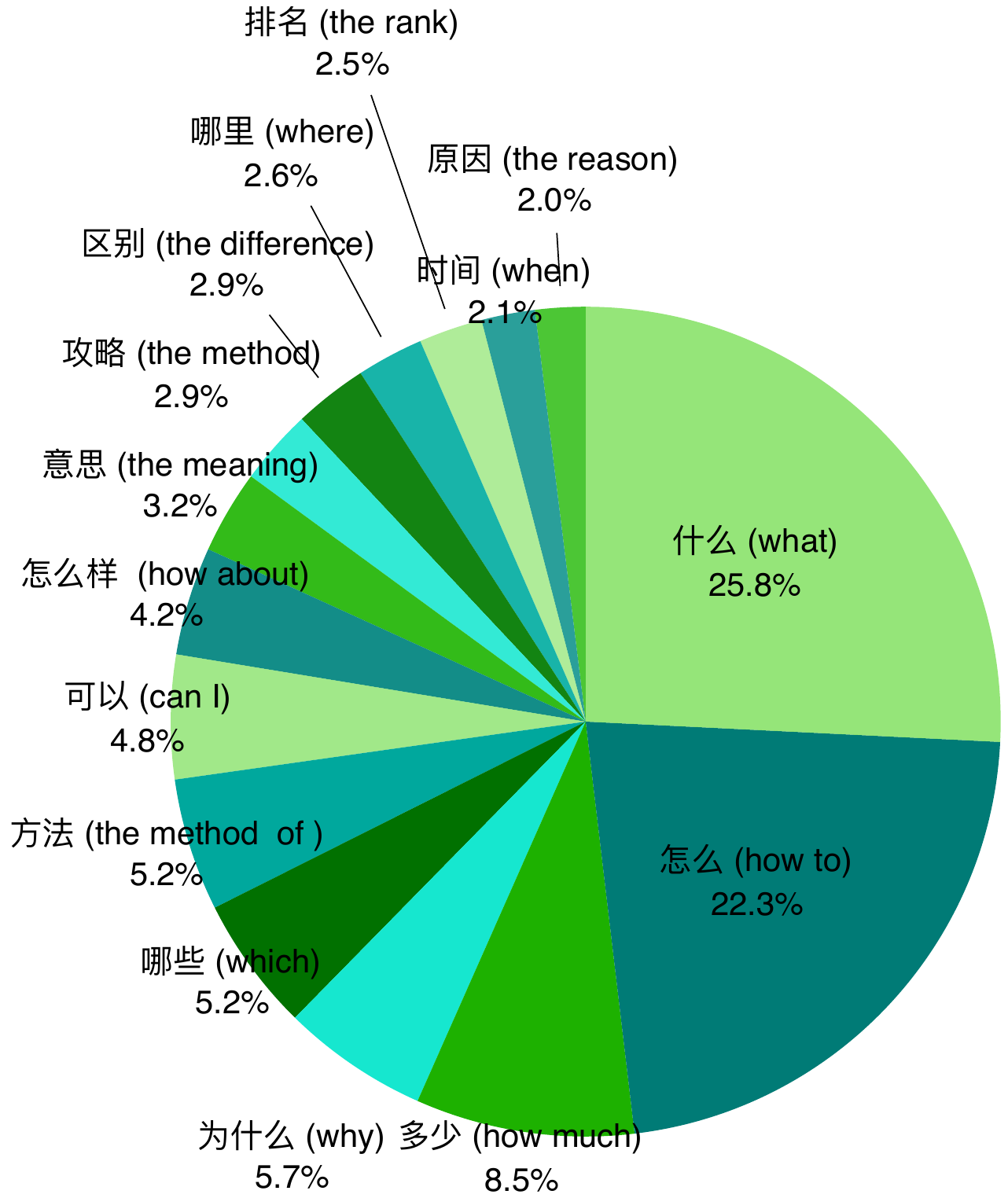}
		\caption{The type of the questions in ComQA.}\label{fig_question_type}
	\end{figure}

	\subsubsection{\textbf{Documents Processing}}
	Now we processed the document into a list of nodes (i.e., sentences and special HTML elements). We first extract the main content of the page. Then we use the hierarchical structure of the HTML to locate the leave nodes in the dom tree. For each leave node, we use its structure features and attributes features to determine whether it is a text node or a special HTML element. We remove the HTML element which is not text content or special HTML element. Finally, we merge the redundant nodes and unnecessary items to form the cleaned document.

	\subsubsection{\textbf{Answer Annotation}}
	We employed crowdworkers to select the answers from the document for each question. We build an HTML based annotation platform so that the crowdworkers could annotate on both computer and mobile devices. For the annotation page, we treat each node in the document as a single element that could be clicked as the answer component. The final answer is the combination of the selected nodes. For the page that the title is not a valid question or the document doesn't contain the right answer, crowdworkers are asked to select none of the nodes. A snapshot of the annotation page is shown in Figure \ref{fig_label}.

	\subsubsection{\textbf{Quality Inspection}}
	We conduct a quality inspection after the data has been annotated by the crowdworkers. Since the dataset is really large, we must \textit{select} the data samples that may be erroneously annotated. We do this by the following stages: first of all, similar with cross-validation \cite{hastie2009elements}, we divided the data into ten folds, we train a model on the nine folds and evaluate on the other one fold. For the one fold held out data, we select the samples that are ambiguous to the current model, i.e. the log probability of the real answer is very small, as the candidate wrong data. This process is conducted ten times, and we select the most unconfident data for the model. This fake data selection process is similar to recent researches which find the hard to learn samples often correspond to labeling errors \cite{swayamdipta2020dataset}. Finally, we feed those samples to the authoritative checkers who are well informed of the ComQA tasks and relabeled those samples.
	
	After the above four processes, we obtain the training ComQA samples that have gone through the quality inspection. Each data instance is made up of: a URL, a question, a document containing a list of quaternion with the form $ <id,type,content,label> $, id is the index of the nodes in the document, type is the node type, for example, sentence or image, label could be 0 or 1 indicating whether the node is an answer component.
	
	\subsubsection{\textbf{Test Set Construction}}
	Since the test set demands a higher quality than the training set, we ask the two authoritative checkers to annotate the same data and select the instances with which the two checkers accord.
	
	Finally, we get 122,343 samples from crowdworkers. We use 5000 of them as the development set and the rest 117,343 as the training set. Moreover, we get 2,054 test samples annotated by authoritative checkers. The documents have average 581 words, and the queries have 7.4 words on average. And each document containing about 23.1 nodes, with 21.2\% (4.9) of them are annotated as the answer nodes.
	Each node has average 24.8 words. The questions in ComQA cover a broad range of topics such as medical, entertainment, etc. The most common question words in ComQA is shown in Figure \ref{fig_question_type}.
	
	\begin{figure*}
		\centering
		\includegraphics[width=1.0\linewidth]{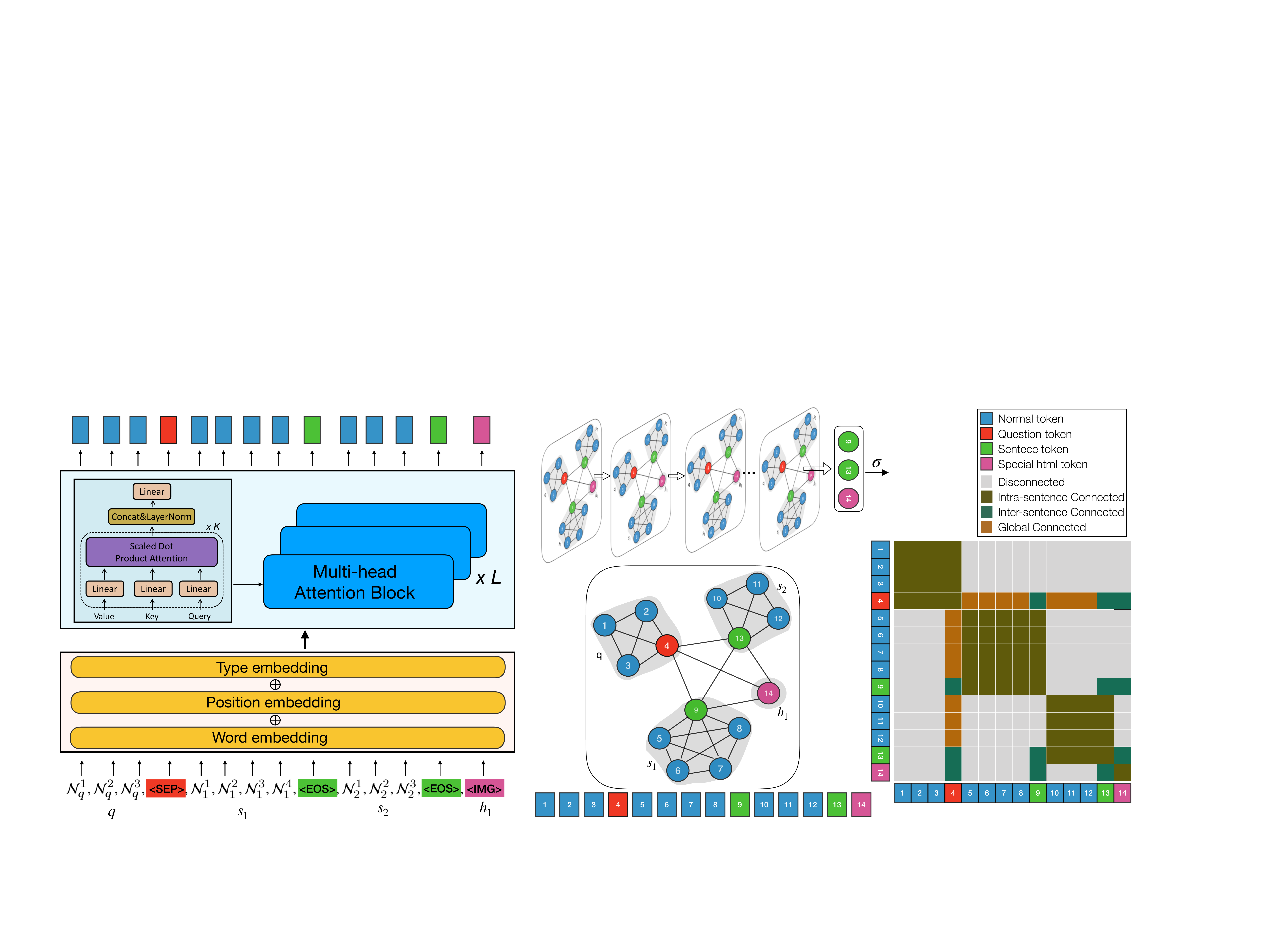}
		\caption{Hierarchical graph neural networks for ComQA, the case shows how to represent the question ($ q $), two sentences ($ s_1,s_2 $) and a special image node ($ h_1 $). \textbf{\textit{Left}}:, the basic BERT sequence encoder to obtain the contextualized representations for each token. The input is the concatenation of question and document tokens. We append some special tokens to indicate the question (\texttt{<SEP>}), sentence (\texttt{<EOS>}) and special html image element \texttt{<IMG>}. \textit{Middle}: The hierarchical graph neural networks blocks, which uses the intra-sentence connection, inter-sentence connection, and global connection (omitted for concise view) to build a hierarchical representation of the document graph. The final prediction is made upon the sentence nodes (green) and special html nodes (purple). \textit{Right}: The connection mask matrix (or better known as the adjacency matrix in graph neural networks) used to connect the different tokens in the graph.
		}\label{fig_model}
	\end{figure*}
	
	%which hinders the standard span extraction techniques that are widely used in machine reading comprehension.

	\section{Hierarchical Graph Neural Networks for ComQA}
	In this section, we describe the proposed hierarchical graph neural networks (HGNN) for the compositional QA. Denote the document as a graph $ \mathcal{G} $, which is a tuple $ (\mathcal{N}, \mathcal{E}) $. $ \mathcal{N}=\{\mathcal{N}_i\}_{i=1}^{N_\mathcal{N}} $ is the node set, and the $ i_{\text{th}} $ node $ \mathcal{N}_i $ consists of a sequence of words $ \{\mathcal{N}_i^{1},\mathcal{N}_i^{2},...,\mathcal{N}_i^{|\mathcal{N}_i|}\} $. $ |\mathcal{N}_i| $ is the number of words in the $ i_{\text{th}} $ node. We represent the question as a special node in $ \mathcal{N}_q $. $ \mathcal{E}=\{e_i\}_{i=1}^{N_\mathcal{E}} $ is the edge set. The prediction is made upon the normal nodes $ \{\mathcal{N}_i\}_{i=1}^{N_\mathcal{N}} $. Our proposed HGNN consists of three modules: the basic sequence encoder, hierarchical graph attention layer, and final prediction layer, the whole architecture is illustrated in Figure \ref{fig_model}.

	\subsection{Sequence Encoder}
	The sequence encoder is a BERT-like module to obtain the basic representations. First of all, we represent the question and document into a single sequence by concatenating the question tokens and every document node tokens. We add a special \texttt{<\textsc{SEP}>} token after the question tokens, and add a special \texttt{<\textsc{EOS}>} token after each sentence. For each special HTML node, we also use the special tokens, such as \texttt{<IMG>}, \texttt{<BR>}, etc., to represent them.
	
	\subsubsection{\textbf{Embedding Layer}}
	We first use the Byte Pair Encoding (BPE) \cite{sennrich2016neural} to tokenize the sequence into word pieces. Suppose the sequence length is $ N $, for each of the word pieces in the sequence, we use the word embedding layer to transform them into $ \textbf{H}_w \in \mathbb{R}^{N \times D} $ where $ D $ is the embedding size. We also use position embedding to get the sequence positional embedding $ \textbf{H}_p \in \mathbb{R}^{N \times D} $. Finally, we also apply the type embedding to the sequence resulting $ \textbf{H}_t \in \mathbb{R}^{N \times D} $ to indicate whether it is in the question or in the document. The final embedding output is the addition of the three embeddings:
	\begin{equation}\label{eq.embedding}
	\textbf{H}_0=\textbf{H}_w+\textbf{H}_p+\textbf{H}_t
	\end{equation}
	\subsubsection{\textbf{Self Attention Layer}}
	After the embedding layer, we apply the self-attention based Transformer \cite{vaswani2017attention} layer to the input. A list of $ L $ Transformer blocks are used to project the input embedding into the contextualized representations. For the $ l_{\text{th}} $ layer in the Transformer, the output could be denoted as:
	\begin{equation}\label{eq.transformer}
	\begin{aligned}
	\textbf{Q}_l,\textbf{K}_l,\textbf{V}_l =& \textbf{W}_q^l\textbf{H}_{l-1},\textbf{W}_k^l\textbf{H}_{l-1},\textbf{W}_v^l\textbf{H}_{l-1}, \\
	\hat{\textbf{H}}_l^k=&\text{Softmax}(\frac{\textbf{Q}_l^{T}\textbf{K}_l}{\sqrt{D}})\textbf{V}_l, \forall k \in[1, K]\\
	\bar{\textbf{H}}_l=&\text{Projection} (\text{Concat}([\hat{\textbf{H}}_l^1,...,\hat{\textbf{H}}_l^K]))\\
	\textbf{H}_l=&\text{LayerNorm} (\bar{\textbf{H}}_l)\\
	\end{aligned}
	\end{equation}
	$ \textbf{W} $ is the learnable weight matrices and the $ \text{Project} $ is a multi-layer perceptron to project the output of the multi-head attention back to the hidden space. $ K $ is the number of \textit{head} in Transformer. Layer normalization \cite{Ba2016LayerN} is applied before the output.

	\subsection{Hierarchical Graph Neural Networks}
	
	In HGNN, node embeddings are initialized with sequence encoder's last layer output $ \textbf{H}_L $, followed by hierarchical graph representations. Formally, the HGNN is built upon the general message passing architecture:
	\begin{equation}\label{eq.gnn}
	\small
	\textbf{H}^{k}=f\left(A, \textbf{H}^{k-1} ; \textbf{W}^{k}\right)
	\end{equation}
	where $ \textbf{H}^{k} \in \mathbb{R}^{N \times D} $ are the node embeddings after $ k $ hop of computation in the graph, $ A \in \mathbb{R}^{N \times N}$ is the adjacency matrix representing the graph structure. $ \textbf{W}^{k} \in \mathbb{R}^{D \times D} $ is the trainable parameters for different graph layers. $ f $ is the message propagation function for information aggregation. In this paper, the layer-wise propagation rule of HGNN is formulated as:
	\begin{equation}\label{eq.agg}
	f\left(A, \textbf{H}^{k-1} ; \textbf{W}^{(k)}\right)=\operatorname{GeLU}\left(\tilde{D}^{-\frac{1}{2}} \tilde{A} \tilde{D}^{-\frac{1}{2}} \textbf{H}^{k-1} \textbf{W}^{k}\right)
	\end{equation}
	where Gelu is the gaussian error linear activation function \cite{hendrycks2016gaussian}. $ \tilde{A}=A+I, \tilde{D}=\sum_{j} \tilde{A}_{i j} $ are used to normalization since directly multiplication with $ A $ will completely change the scale of the input. There are $ M $ layers (hops) in the HGNN and the final output is $ \textbf{H}^M $.
	
	\subsubsection{\textbf{Graph Construction}} \label{sec.graph.construct}
	$ \\ $
	Graph construction is one of the most important factors for the good performance of graph neural networks \cite{hamilton2017representation,kipf2016semi}. In this paper, we consider three types of connection:
	
	\textbf{Intra-Sentence Connection}: For each word in a sentence, we connected them with each other. Specifically, the special tokens such as $ \texttt{<\textsc{EOS}>} $ are also connected with other words in the same sentence:
	\begin{equation*}
	A_{i,j}^{\text{intra}}=\left\{\begin{array}{ll}
	1 & \text {\textit{if i and j are located in the same sentence.}} \\
	0 & \text {\textit{otherwise.} }
	\end{array}\right.
	\end{equation*}
	Thus, the intra-sentence graph connection is formulated by the adjacent matrix $ A^{\text{intra}} $.
	
	\textbf{Inter-Sentence Connection}: We have already appended the special tokens after each sentence as the high-level node identification, so we add the inter-sentence connection to those special tokens in the document. If the node corresponding tokens belongs to the special tokens, i.e. \texttt{<\textsc{SEP}>}, \texttt{<\textsc{EOS}>} and the special HTML element tokens, we add a connection between them:
	\begin{equation*}
	A_{i,j}^{\text{inter}}=\left\{\begin{array}{ll}
	1 & \text {\textit{if i and j belongs to the special tokens}} \\
	& \text {\textit{indicating the higher level node.}} \\
	0 & \text {\textit{otherwise}.}
	\end{array}\right.
	\end{equation*}
	
	\textbf{Global Connection}: When represent the document in QA, question attention is very important \cite{Seo2016BidirectionalAF,yu2018qanet}. However, only employing the inter-sentence connection, where the interaction between the words in the document and the words in the question are all based on their high-level sentence indicators, may incur the modeling burden for the attention based model \cite{zaheer2020big}. Therefore, we also construct the global connection between the question indicator \texttt{<\textsc{SEP}>} and all other words in the document:
	\begin{equation*}
	A_{i,j}^{\text{global}}=\left\{\begin{array}{ll}
	1 & \text {\textit{if i is \texttt{<\textsc{SEP}>}},} \forall j \in[1, N] \\
	0 & \text {\textit{otherwise}.}
	\end{array}\right.
	\end{equation*}
	
	\subsubsection{\textbf{Information Aggregation}} 
	$ \\ $
	After the graph construction process, we build the graph representations based on Equation \ref{eq.agg}. However, since the graph is hierarchical containing different level of nodes with different connections, we have two options to represent them:
	
	\textbf{Pipeline Aggregation}: We could build the hierarchical representation in pipeline, that is, we first build the low-level intra-sentence representations, then we build the higher level inter-sentence and global representations:
	\begin{equation}\label{eq.pipe}
	\begin{aligned}
	&\textbf{H}_{\text{intra}}^k=\operatorname{GeLU}\left(\tilde{D}_{\text{intra}}^{-\frac{1}{2}} \tilde{A}_{\text{intra}} \tilde{D}_{\text{intra}}^{-\frac{1}{2}} \textbf{H}^{k-1} \textbf{W}_{\text{intra}}^{k}\right)\\
	&\textbf{H}_{\text{inter}}^k=\operatorname{GeLU}\left(\tilde{D}_{\text{inter}}^{-\frac{1}{2}} \tilde{A}_{\text{inter}} \tilde{D}_{\text{inter}}^{-\frac{1}{2}} \textbf{H}_{\text{intra}}^k \textbf{W}_{\text{inter}}^{k}\right)\\
	&\textbf{H}^k=\operatorname{GeLU}\left(\tilde{D}_{\text{global}}^{-\frac{1}{2}} \tilde{A}_{\text{global}} \tilde{D}_{\text{global}}^{-\frac{1}{2}} \textbf{H}_{\text{inter}}^k \textbf{W}_{\text{global}}^{k}\right)
	\end{aligned}
	\end{equation}

	\textbf{Fused Aggregation}: The pipeline aggregation build the graph representation in a hierarchical way, however, we can instead pack them into a single operation by fusing the three-level adjacency matrices into one:
	\begin{equation}\label{eq.fused}
	A_{i,j}=\left\{\begin{array}{ll}
	1 &  \text{if }{ \left\{\begin{array}{ll}
		A_{i,j}^{\text{intra}} = 1 \text{ or}\\
		A_{i,j}^{\text{inter}} = 1 \text{ or}\\
		A_{i,j}^{\text{global}}= 1 
		\end{array}\right.} \\
	0 & \text {\textit{otherwise}.}
	\end{array}\right.
	\end{equation}
	
	Based on the fused adjacency matrix, we can build the graph representations based on Equation \ref{eq.agg}. The adjacency matrix is also illustrated in the right of Figure \ref{fig_model}. In experiment, we will compare the two types of aggregation scheme.
	
	\subsection{Prediction and Objective}\label{sec.predict}
	After representing the question and document by the graph neural networks, we collect the sentence-level nodes and make predictions on them. Those sentence level nodes including the \texttt{<\textsc{EOS}>} and special HTML tokens, forming the answer candidate set $ \mathcal{S} $. The final probability of each node in  $ \mathcal{S} $ is calculated by:
	\begin{equation}\label{eq.prob}
	P_{i \in  \mathcal{S}} = \sigma(\textbf{H}^M_{i \in  \mathcal{S}}\textbf{w}_o)
	\end{equation}
	where $ \sigma $ is the sigmoid function and $ \textbf{w}_o $ is the weight vector to project the last layer representations of HGNN into scalar values.
	
	The objective of the model is to minimize the binary cross-entropy between the predictions and real labels:
	\begin{equation}\label{eq.loss}
	\mathcal{L}_{BCE}=-\mathbbm{1}_{i \in \mathcal{S}_P} \log \left(P_{i \in  \mathcal{S}_P} \right)  -\mathbbm{1}_{i \in \mathcal{S}_N} \log \left(1-P_{i \in  \mathcal{S}_N} \right) 
	\end{equation}
	where $ \mathcal{S}_P $ is the positive (1) nodes set and $ \mathcal{S}_N $ is the negative (0) nodes set in a document, $ \mathcal{S}_P \cup \mathcal{S}_N=\mathcal{S} $.

	\subsection{Unsupervised Pre-training}\label{sec.pretrain}
	Recent works on NLP have shown the great advantage of large-scale pre-training \cite{devlin2018bert,radford2019language,roberts2020much}. However, different from the previous works of pre-training where the input is the raw text sequence, the ComQA contains the information of document structure. Therefore, we devised three types of unsupervised objectives for pre-training:
	
	\textbf{Masked Language Model} (\textbf{MLM}): similar to \citet{devlin2018bert}, we masked some words in the sequence, and the model must predict the original words based on the surrounding context. We mask 15\% of the sequence and do not mask the special tokens.
	
	\textbf{Question Selection} (\textbf{QS}): Since ComQA is constructed from the web pages where the question is the page's title and the document is the page content. There is an inherent correlation between the question and the document. So we propose the question selection pre-training, a binary classification task to determine whether the question, i.e., the title, is relevant to the document.
	We replace the title with a random title as the \textit{fake} example, and the original title-document as the \textit{positive} example. We use the representations of the question indicator $ \textbf{H}^{M}_{i=\texttt{<\textsc{SEP}>}} $ for prediction just as in Equation \ref{eq.prob}, except that we use the different weight vector.
	
	\textbf{Node Selection} (\textbf{NS}): In addition to the question selection, we can also perform a node selection task on the nodes to determine whether it is relevant to its contextual document. This task is similar to the ComQA where the prediction is also made on document nodes. We use two heuristic rules to construct the negative nodes: 1) we replace the sentence in the document with a random sentence. 2) we randomly shuffle the words in the sentence and treat the resulting sentence as a negative node. 3) we randomly swap the two nodes in the document and treat the two nodes as negative. We also replace 15\% of the original nodes in the document as the negative nodes.

	\renewcommand\arraystretch{1.06} 
	\begin{table*}[]
%		\small
		\centering
		\setlength{\tabcolsep}{3.8pt}{
			\begin{tabular}{clcccccccc}
				& & \multicolumn{4}{c}{\textit{Dev}}                                   & \multicolumn{4}{c}{\textit{Test}}             \\ \cline{3-10} 
				& & Precision & Recall & F1   & \multicolumn{1}{c|}{Accuracy} & Precision & Recall & F1   & Accuracy \\ \noalign{\hrule height 1.15pt}
				&\multicolumn{1}{l|}{SequentialTag$ _{\text{lstm}} $} & 51.7      & 59.1   & 56.0 & \multicolumn{1}{c|}{33.9}     & 43.2      & 56.1   & 47.9 & 25.3     \\
				& \multicolumn{1}{l|}{SequentialTag$ _{\text{bert}} $} & 65.3      & 78.2   & 73.1 & \multicolumn{1}{c|}{38.2}     & 61.9      & 73.7   & 66.5 & 32.9     \\ \hline \hline
				& \multicolumn{1}{l|}{LSTM}                & 73.1      & 39.9   & 55.3 & \multicolumn{1}{c|}{28.2}    & 83.4      & 46.1   & 59.4 & 28.2    \\
				& \multicolumn{1}{l|}{BiDAF}               & 72.5      & 64.7   & 69.1 & \multicolumn{1}{c|}{35.7}     & 69.4      & 58.7   & 63.9 & 30.2     \\
				& \multicolumn{1}{l|}{QA-Net}              & 75.4      & 63.3   & 68.8 & \multicolumn{1}{c|}{36.6}     & 71.6      & 60.9   & 65.3 & 31.8     \\
				& \multicolumn{1}{l|}{BERT$ _{\text{official}}^* $}       & 81.9      & 64.8   & 70.9 & \multicolumn{1}{c|}{35.3}    & 81.1      & 66.2 & 72.9 & 35.4  \\
				& \multicolumn{1}{l|}{BERT$ _{\text{base}} $}           & 84.2      & 78.3   & 80.5 & \multicolumn{1}{c|}{42.9}    & 73.5      & 75.4   & 74.1 & 37.9    \\
				& \multicolumn{1}{l|}{BERT$ _{\text{large}} $}         & 85.1      & 79.2   & 82.1 & \multicolumn{1}{c|}{46.3}    & 77.3      & 78.0   & 77.6 & 43.2  \\ \hline \hline
				\multirow{4}{*}{\rotatebox{90}{\parbox{1.5cm}{\centering Pipeline}}}& \multicolumn{1}{l|}{HGNN$ _{\text{small}} $}          & 79.3      & 76.1   & 76.9 & \multicolumn{1}{c|}{42.0}    & 68.2      & 74.5   & 72.0 & 36.5      \\
				& \multicolumn{1}{l|}{HGNN$ _{\text{base}} $}          & 83.0      & 81.7   & 81.9 & \multicolumn{1}{c|}{44.8}    & 74.2      & 75.9   & 75.1 & 40.7    \\
				& \multicolumn{1}{l|}{HGNN$ _{\text{large}} $}         & 84.4      & 82.5   & 83.3 & \multicolumn{1}{c|}{48.2}     & 77.1      & 78.9   & 78.0 & 44.9       \\ 
				& \multicolumn{1}{l|}{HGNN$ _{\text{large}} $+QS+NS}          & 85.9      & 83.1   & 84.2 & \multicolumn{1}{c|}{50.8}      & 78.4      & 78.5   & 78.4 & 46.3     \\
				\hline 
				\multirow{4}{*}{\rotatebox{90}{\parbox{1.4cm}{\centering \textsc{Fused}}}}& \multicolumn{1}{l|}{HGNN$ _{\text{small}} $}          & 80.2      & 75.1   & 77.2 & \multicolumn{1}{c|}{42.9}    & 70.7      & 75.8   & 73.9 & 38.2      \\
				& \multicolumn{1}{l|}{HGNN$ _{\text{base}} $}          & 86.0      & 82.1   & 83.4 & \multicolumn{1}{c|}{47.3}    & 76.6      & 78.1   & 77.3 & 43.8    \\
				& \multicolumn{1}{l|}{HGNN$ _{\text{large}} $}         & 86.9      & 83.7   & 85.4 & \multicolumn{1}{c|}{50.7}     & 78.9      & 80.0   & 79.0 & 47.1       \\ 
				& \multicolumn{1}{l|}{HGNN$ _{\text{large}} $+QS+NS}          & \textbf{87.3}      & \textbf{84.9}   & \textbf{86.8} & \multicolumn{1}{c|}{\textbf{53.9}}      & \textbf{79.6}      & \textbf{80.3}   & \textbf{79.9} & \textbf{49.2}     \\
				\hline \hline
				& \multicolumn{1}{l|}{Human}         & -      & -   & - & \multicolumn{1}{c|}{-}     & 88.5      & 91.0   & 89.8 & 80.5     \\ \noalign{\hrule height 1.15pt}
		\end{tabular}}
	\caption{Main results on ComQA. The first SequentialTag model is based on sequential tagging to select the relevant segments in the doc. The middle six models are based on the plain sequential tags without structure information. For BERT$ _{\text{official}} $ model, since the max sequence length was limited to 512, the recall and accuracy are relatively low. Pipeline and Fused refers to the two graph information aggregation operations. QS and NS refers to our model pre-trained with additional question selection and node selection tasks described in Section \ref{sec.pretrain}.}\label{tab.result}
	\end{table*}

	\section{Experiments}
	\subsection{Common Setup}
	In all experiments, we tokenize the text with BPE by sentencepiece \cite{kudo2018sentencepiece}. We set the vocabulary size to 50,000. We use the Adam \cite{Kingma2014AdamAM} optimizer with 5k warm-up steps and linearly decay the learning rate. $ \beta_1, \beta_2, \epsilon $ was set to 0.9, 0.99 and $ 10^{-6} $, respectively. For both pre-training and fine-tuning, the max learning rate was set to $ 10^{-4} $. We applied decoupled weight decay \cite{loshchilov2018decoupled} to the model with scale 0.1. Dropout \cite{srivastava2014dropout} was applied to every parameter and attention mask with drop probability 0.1. We clip the max grad norm to 0.2.
	The batch size was 256 during pre-training and 64 during fine-tuning. We truncate the document sequence length to 1024. We use Pytorch \cite{paszke2019pytorch} 1.4.0 framework. All experiments are conducted through 16 volta V100 GPUs.
	
	We trained three types of HGNN model. HGNN$ _{\text{small}}$ is the small version of the model containing 6 layers of the Transformer encoder, each layer contains 8 heads with hidden size 512. HGNN$ _{\text{base}}$ is the base version of the proposed model, which contains 12 layers of the Transformer encoder, each layer containing 12 heads with hidden size 768. HGNN$ _{\text{large}}$ is the large version of the proposed model containing 24 layers of Transformer encoder, the head, and hidden size was set to 16, 1024, respectively. For the hierarchical graph neural networks, we set the number of layers $ M $ to 4. 
	
\begin{figure}[!ht]
		\centering
		\includegraphics[width=0.8\linewidth]{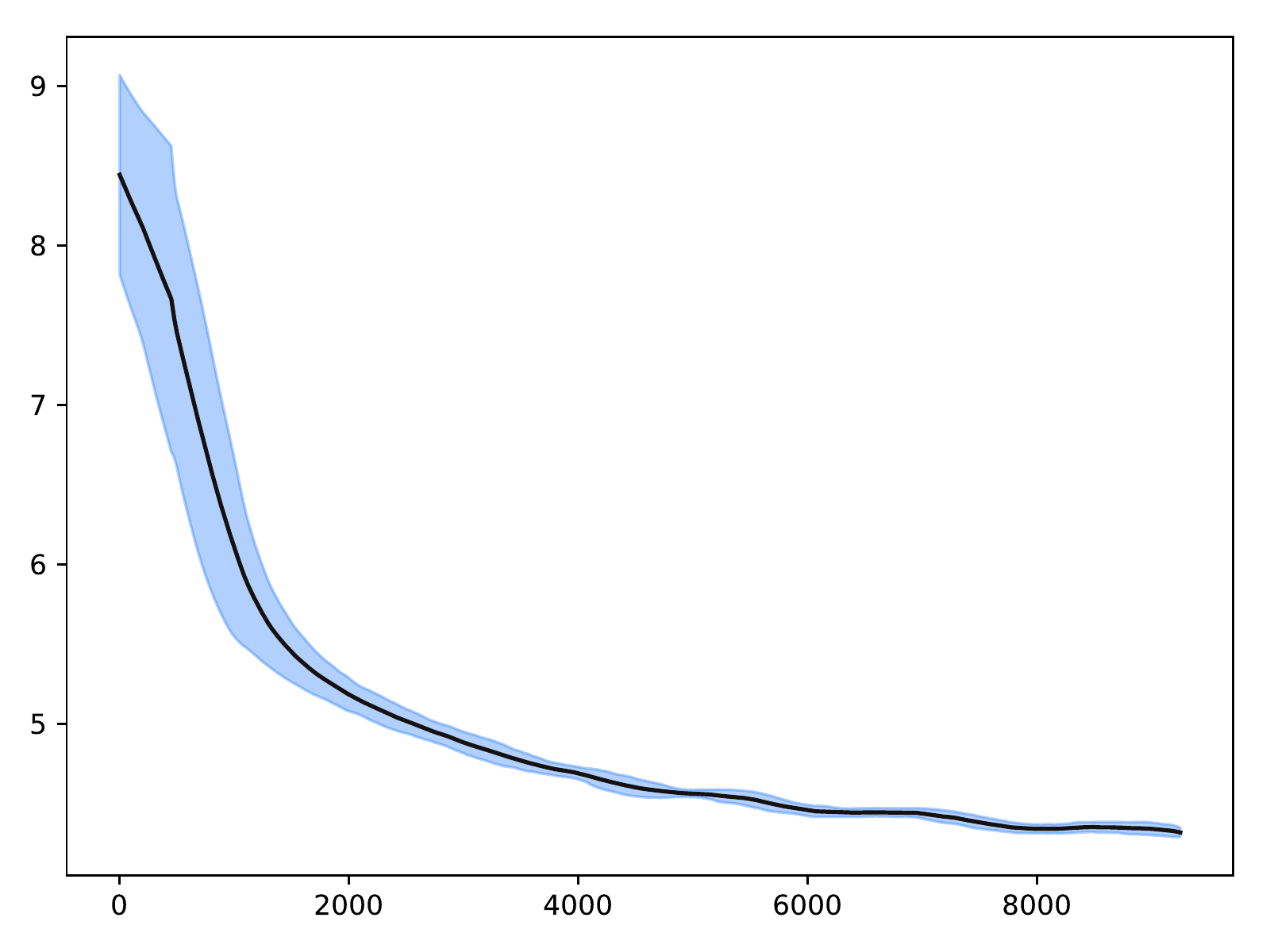}
		\caption{The pre-training loss of the proposed HGNN$ _{\text{base}} $ model. The x-axis is the pre-training steps ($ \times 100 $).}\label{fig.pre.train.loss}
	\end{figure}

	We use 400 million Chinese web pages for pre-training, those data occupy nearly 1.2Tb disk space. We clean the document by removing the hyperlink, phone number, or special symbols. It needs to mention that for the QS pre-training task, we restrict the title to be a valid question. For other types of pre-training such as MLM and NS, we just use the unfiltered web pages. We conduct 1 million pre-training steps for all pre-training model. The pre-training loss of HGNN$ _{\text{base}}$ is shown in Figure \ref{fig.pre.train.loss}.
	
	\subsection{Evaluation Metrics}
	In ComQA, each document has several nodes that could be selected as the answer components, denote the nodes that the model predict is true as $ \mathcal{S}_R:\{i|P_i > \text{threshold} \} $. For each document, we could define the \textit{Precision}, \textit{Recall} and \textit{F1} and \textit{Accuracy}:
	\begin{equation*}
	\begin{aligned}
	&\text{Precision}=\frac{|\mathcal{S}_P \bigcap \mathcal{S}_R|}{|\mathcal{S}_P|}\\
	&\text{Recall}=\frac{|\mathcal{S}_P \bigcap \mathcal{S}_R|}{|\mathcal{S}_R|}\\
	&\text{F1}=\frac{2 \times \text{Precison}\times \text{Recall} }{\text{Precison}+\text{Recall}}\\
	&\text{Accuracy}={ \left\{\begin{array}{ll}
		1 & \text{if  }\mathcal{S}_P \equiv \mathcal{S}_R\\
		0 & \text{otherwise}
		\end{array}\right.}
	\end{aligned}
	\end{equation*}
	The threshold was tuned in the development set. The accuracy is a binary value denoting whether all answer nodes are correctly selected. And we average every document's precision, recall, F1, and accuracy as the final precision, recall, F1, and accuracy, respectively. It needs to mention that the positive node set $ \mathcal{S}_P $ could be empty.

	\subsection{Baseline Methods}
	We compare our model with five baseline models:
	\begin{itemize}[leftmargin=*]
		\setlength{\itemsep}{0.1cm}
		\setlength{\parsep}{0.0cm}
		\setlength{\parskip}{0.0cm}
		\item \textbf{SequentialTagging}: We use a sequential tagging method which predicts the \textit{begin
		}, \textit{in} and \textit{out} (BIO) tag of each word in the document. This model is flexible to select the discrete segments in the document. However, the selected segments are not guaranteed to match a sentence. Therefore, we select the sentence with more than half of its words are annotated as $ B $ or $ I $, as the prediction set $ \mathcal{S}_R $. Specifically, we select two types of sequential tagging model, the first one is based on BiLSTM \cite{chiu2016named}, and the second one is based on BERT \cite{devlin2018bert}.
		
		\item \textbf{LSTM} \cite{hochreiter1997long} is based on the bi-directional LSTM to process the question and document in a sequential way. Like the proposed HGNN, we concatenate the question and document tokens. The prediction is also made on the special token. 
		
		\item \textbf{BiDAF} \cite{Seo2016BidirectionalAF} is a widely used attention-based method. It builds the LSTM based bi-directional attention flow between the words in the question and the words in the documents, which achieves promising results in many QA tasks. Other than predicting the start and end position of the answer span, we use the same prediction methods in Section \ref{sec.predict} for answer nodes selection.
		
		\item \textbf{QA-Net} \cite{yu2018qanet} is the current state-of-the-arts machine reading comprehension method without pre-training. QAnet does not require recurrent networks but consists exclusively of convolution and self-attention, where convolution models local interactions and
		self-attention models global interactions. Different from the original paper \cite{yu2018qanet}, we only adopt the architecture and did not apply backtranslation for data augmentation.
		
		\item \textbf{BERT} \cite{devlin2018bert} is a pre-training method that pushes many NLP tasks into new record. It is based on the encoder of the Transformer \cite{vaswani2017attention}, the objective is the masked language modeling and next sentence prediction. In fact, our model without hierarchical graph neural networks is reduced to the BERT model. We use two types of BERT model, the first one is the official released Chinese version BERT base model\footnote{\url
			https://github.com/huggingface/transformers} (BERT$ _\text{official} $). However, the max length of the official BERT is 512, which will truncate too many samples in ComQA. So we also pre-train the BERT on the same data we pre-trained the HGNN with max length 1024. We also train two types of BERT model, i.e. BERT$ _\text{base} $ and BERT$ _\text{large} $.
	\end{itemize}
	
	\renewcommand\arraystretch{1.2} 
	\begin{table}
		\small
		\centering
		\setlength{\tabcolsep}{3.5pt}{
			\begin{tabular}{lccccc}
				& \multicolumn{1}{l}{\#Head} & \multicolumn{1}{l}{\#Layer} & \multicolumn{1}{l}{\#Hidden} & \multicolumn{1}{l}{\#Interconnect} & \multicolumn{1}{l}{\#Parameter} \\ \noalign{\hrule height 1.15pt}
				LSTM  & -                          & 3                           & 128                           & -                              & 6,796,544                     \\ \hline
				QA-Net  & -                          & -                           & 128                           & -                              & 13,876,706                     \\ \hline
				BiDAF  & -                          & -                           & 100                           & -                              & 10,747,168                     \\ \hline
				BERT$ _{\text{official}} $  & 12                          & 12                           & 768                           & 3072                               & 102,268,416                     \\ \hline
				BERT$ _{\text{base}} $      & 12                          & 12                           & 768                           & 3072                               & 102,268,416                     \\ \hline
				BERT$ _{\text{large}} $     & 16                          & 24                           & 1024                          & 4096                               & 329,486,296                     \\ \hline
				HGNN$ _{\text{small}} $    & 8                           & 6                            & 512                           & 2048                               & 45,561,856                      \\ \hline
				HGNN$ _{\text{base}} $     & 12                          & 12                           & 768                           & 3072                               & 125,400,576                     \\ \hline
				HGNN$ _{\text{large}} $   & 16                          & 24                           & 1024                          & 4096                               & 362,579,328  \\ \noalign{\hrule height 1.15pt}
		\end{tabular}}
		\caption{The specific configuration of different models.}\label{tab.configuration}
	\end{table}
	
	The specific configurations of those models are listed in Table \ref{tab.configuration}. In addition to the above baselines, we also evaluate human performance. We randomly sample 200 questions from the test sets and asks a volunteer to get human performance. The result of those models in dev and test set are shown in Table \ref{tab.result}.

	\subsection{Main Results}
	We can see from the table that our model achieves the best result compared with the other baseline models. For the sequential tagging model, the prediction is made upon the word level, but the evaluation is made upon the sentence level. This training-evaluation gap results in poor performance in ComQA. The performance of LSTM is really poor since it contains no \textit{attention} mechanism that is vital for QA \cite{huang2018fusionnet}. Besides, the input of the ComQA is a document containing more than 500 tokens on average. The current sequential model may have difficulty processing such long sequence.
	For BiDAF and QA-Net, although they have achieved good performance in many QA tasks, they still lag behind the pre-training methods. 
	
	The proposed HGNN is based on BERT, nevertheless, it achieves 3.2/5.9 absolute gain of F1/accuracy on the based version of the model, and 1.4/3.9 absolute improvement on the large model, which shows the effectiveness of the proposed hierarchical graph neural networks. The HGNN is based on BERT, but it introduces the inductive bias of the document structure to the model. From our point of view, the graph structure guides the information aggregation from the words to the sentences to the document, which is very important for ComQA.
	
	For the two types of information aggregation methods, we found the fused aggregation, which merges the three-level connection in a single adjacency matrix, has a better performance. Since the pipeline aggregation directly separates the three-level aggregation, the low-level information still has the difficulty to propagate to the high level, let alone the HGNN is multi-layer.
	
	However, as human achieves 80.5 accuracy on the test set, our results suggest that there is still room for improvement in ComQA.
	
	\begin{figure}
		\centering
		\includegraphics[width=1.0\linewidth]{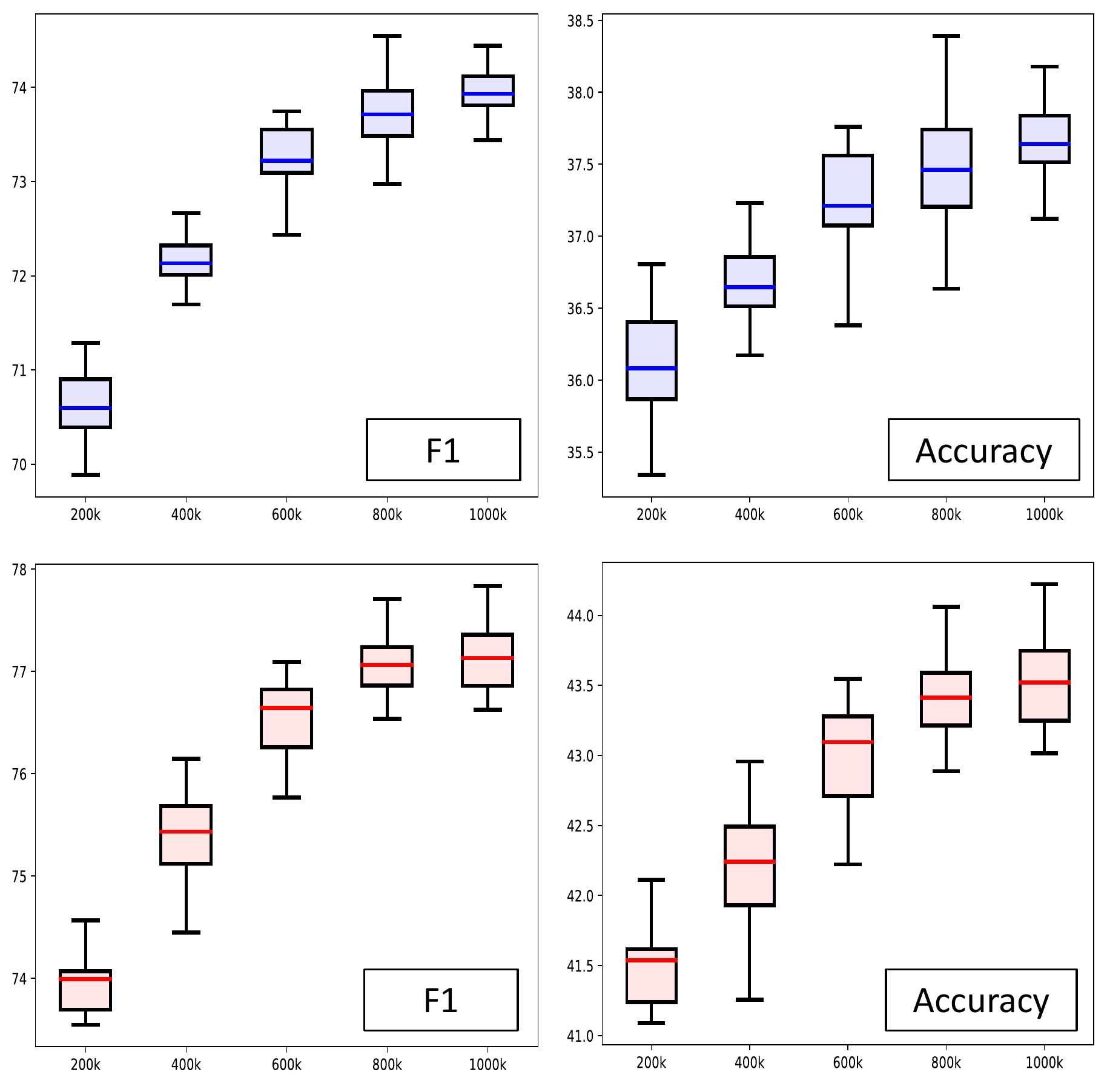}
		\caption{Ablation over number of pre-training steps. The x-axis is the number of parameter updates. We fine-tune each checkpoint 20 epochs, and show the boxplot of the 20 results. We report the F1 and accuracy over the HGNN$ _{\text{base}} $ model (based on Fused connection, upper half) and BERT$ _{\text{base}} $ model (lower half).}\label{fig_pretrain}
	\end{figure}

	\subsection{The Effect of Pre-training}
	We can see from the table \ref{tab.result} that the model with pre-training could obtain significant improvement. In this section, we want to investigate the effect of pre-training in more detail. Concretely, we want to investigate the pre-training in two aspects: the scale of the pre-training and the objective of pre-training.
	
	First of all, \citet{liu2019roberta} and \citet{devlin2018bert} have shown that more pre-training steps could consistently improve the performance of the downstream tasks a lot. So we load the checkpoints of the model during the pre-training dynamics and fine-tuned it on the ComQA dataset. Each model is fine-tuned 20 epochs. We plot the F1 and accuracy of different pre-training model in Figure \ref{fig_pretrain}. We can see that with the pre-training process, the pre-training model could be consistently improved regardless of its architecture. Therefore, scaling is a sure-fire approach for better model quality in ComQA. Nonetheless, when we have pre-trained sufficient number of steps, the performance of the model is saturated. In fact, we find the best results were obtained after nearly 1.2 million pre-training steps, which takes only 25\% of the pre-training data.
	\begin{figure}
		\centering
		\includegraphics[width=1.0\linewidth]{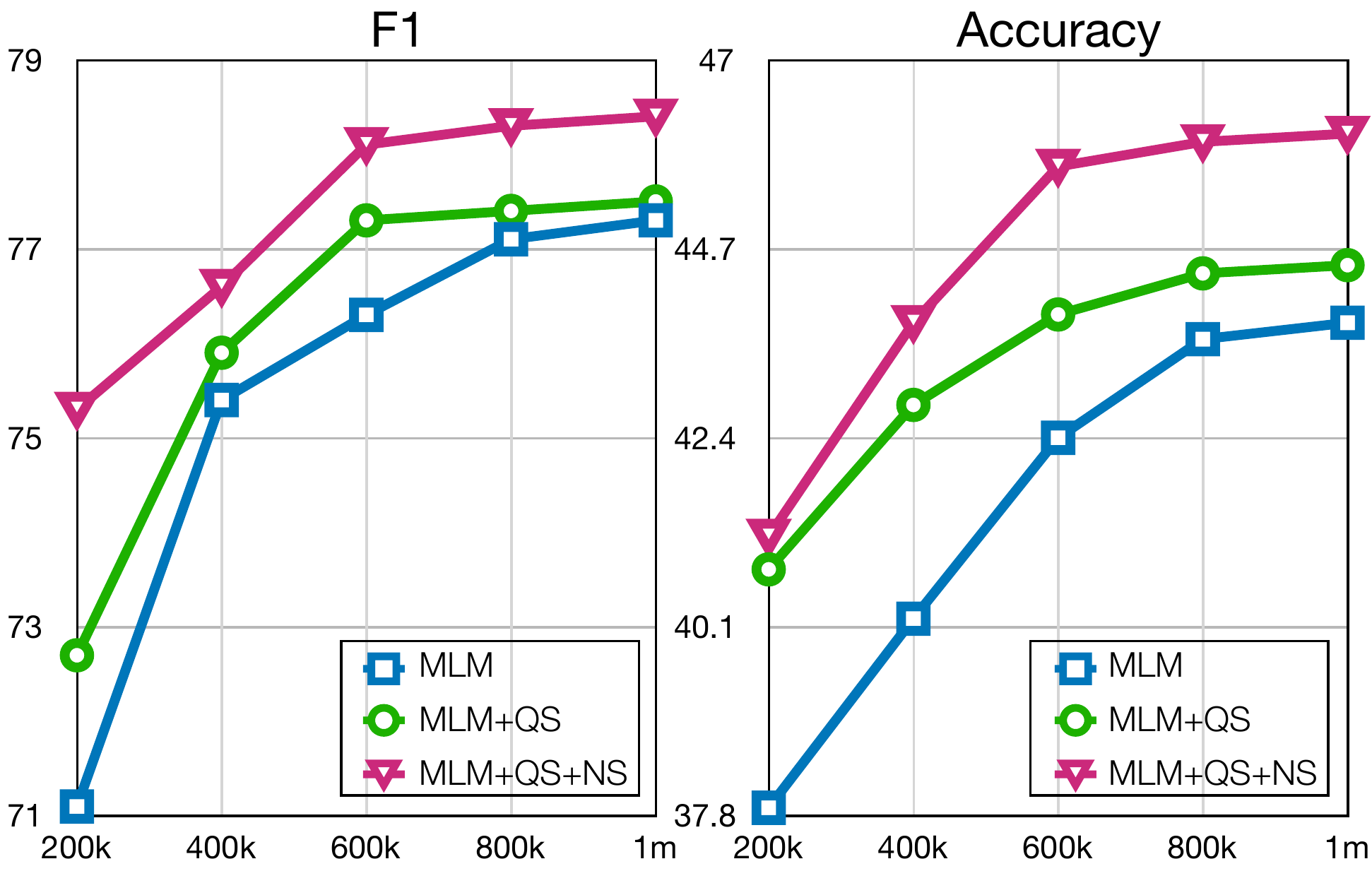}
		\caption{Ablation over the pre-training objectives. The result is based on HGNN$ _{\text{base}} $.}\label{fig_pretraintask}
	\end{figure}
	
	\begin{figure}
		\centering
		\includegraphics[width=0.9\linewidth]{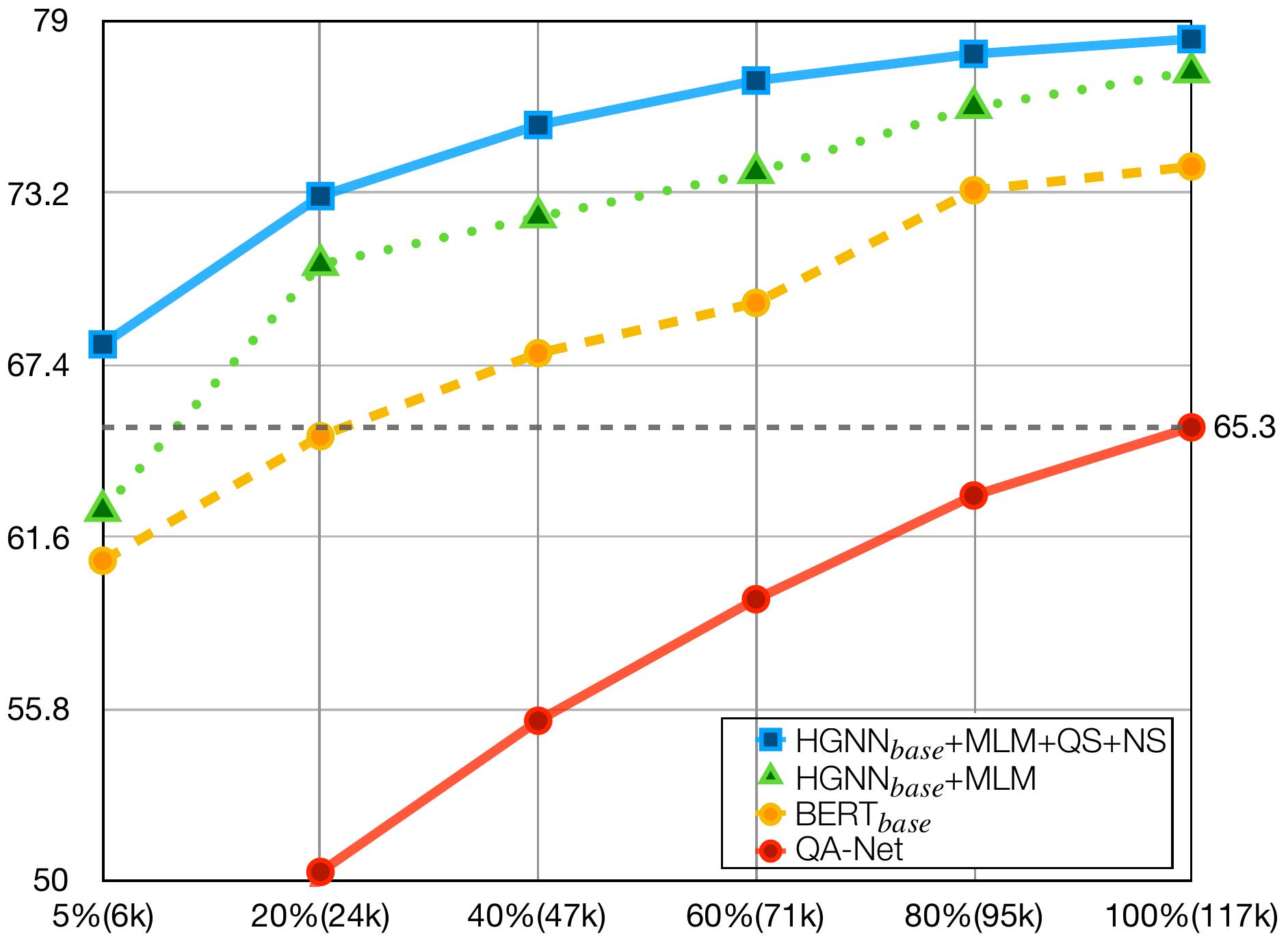}
		\caption{The F1 of the models with different number of fine-tuning ComQA data. The x-axis is the proportion(number) of the fine tuning ComQA data.}\label{fig_pretraindata}
	\end{figure}
	
	Next, we investigate the pre-training tasks proposed in Section \ref{sec.pretrain}. We have already shown in Table \ref{tab.result} that our model could obtain better performance when pre-trained with the additional QS and NS. We conduct the ablation study on the three pre-training objectives for HGNN$ _{\text{base}} $. The result is shown in Figure \ref{fig_pretraintask}.
	
	We can see that the model could achieve a better result in terms of both F1 and accuracy with QS and NS. In the MLM objective, the input is always the corrupted sentence containing the special \texttt{<\textsc{MASK}>} token, which is not the case during fine-tuning. This discrepancy between pre-training and fine-tuning may hurt the model \cite{yang2019xlnet}. On the contrary, our proposed question selection and node selection pre-training tasks are in accordance with the ComQA, where the QS aims at learning the relevance between the question and the document while the NS aims at learning the dependency of a node to a document. In fact, when pre-trained with the NS, our model could be directly applied to ComQA without any new parameters, which results in a faster convergence rate. Figure \ref{fig_pretraindata} shows the performance of the model with the different number of ComQA training data. We can see that the model pre-trained with QS and NS could achieve very good result even with moderate six thousands of labeled data, which excels the full-trained QA-Net. Therefore, devising appropriate pre-training tasks could potentially improve the generalization ability of the model.

	\subsection{The Effect of Connection Type}
	In section \ref{sec.graph.construct}, we introduce three types of the graph connection, i.e. the intra-sentence, inter-sentence, and global connection. As the graph structure of the document is a key factor in ComQA, we analyze their specific importance. Concretely, we build the adjacency matrix based on fused aggregation in Equation \ref{eq.fused}, with the combination of specific connections. 
	
	\renewcommand\arraystretch{1.25} 
	\begin{table}[]
		\centering
		\large
		\begin{tabular}{ccc|cccc}
			\huge{$\clubsuit$} & \huge{$ \heartsuit $} & \huge{$\spadesuit$} & \multicolumn{1}{c}{Precision} & \multicolumn{1}{c}{Recall} & \multicolumn{1}{c}{F1}   & \multicolumn{1}{c}{Accuracy}\\ \hline \hline
			$\surd$  &$\surd$ &$\surd$  & \multicolumn{1}{c}{76.6}      & \multicolumn{1}{c}{78.1}   & \multicolumn{1}{c}{77.3} & \multicolumn{1}{c}{43.8}   \\ \hline
			$\surd$  &  $ \times $    &  $ \times $ & \cellcolor[HTML]{A9A9A9}-2.4      & \cellcolor[HTML]{A9A9A9}-2.2   & \cellcolor[HTML]{A9A9A9}-2.3 & \cellcolor[HTML]{A9A9A9}-5.6     \\
			$ \times $&$\surd$ &  $ \times $  & \cellcolor[HTML]{D9D9D9}-0.6      & \cellcolor[HTML]{D9D9D9}-0.7   & \cellcolor[HTML]{D9D9D9}-0.6 & \cellcolor[HTML]{D9D9D9}-0.8     \\
			$ \times $&$ \times $&$\surd$ & \cellcolor[HTML]{C9C9C9}-1.1      & \cellcolor[HTML]{C9C9C9}-1.8   & \cellcolor[HTML]{C9C9C9}-1.5 & \cellcolor[HTML]{C9C9C9}-3.3     \\
			$\surd$  &$\surd$ &  $ \times $  & \cellcolor[HTML]{E9E9E9}-0.5      & \cellcolor[HTML]{E9E9E9}-0.8   & \cellcolor[HTML]{E9E9E9}-0.6 & \cellcolor[HTML]{E9E9E9}-0.9     \\
			$\surd$  &  $ \times $  &$\surd$ & \cellcolor[HTML]{B9B9B9}-1.3   & \cellcolor[HTML]{B9B9B9}-1.9   & \cellcolor[HTML]{B9B9B9}-1.6 & \cellcolor[HTML]{B9B9B9}-3.5     \\
			$ \times $&$\surd$ &$\surd$ & \cellcolor[HTML]{F9F9F9}-0.3      & \cellcolor[HTML]{F9F9F9}-0.4   & \cellcolor[HTML]{F9F9F9}-0.3 & \cellcolor[HTML]{F9F9F9}-0.6     \\ \hline
			$ \times $&  $ \times $  &  $ \times $  & -3.1      & -2.7   & -3.0 & -5.9    \\ \noalign{\hrule height 1.15pt}
		\end{tabular}
		\caption{Ablation results of the different coonection types of HGNN$ _{\text{base}} $ by fused operation. $\clubsuit$ denotes the intra-sentence connection, $\heartsuit $ denotes the inter-sentence connection and $\spadesuit$ denotes the global connection. $\surd$ means the connection is kept in Equation \ref{eq.fused}. The last raw with no connection reduced to the BERT$ _{\text{base}} $ model.}\label{table.adj.type}
	\end{table}
	
	The result of the test set is shown in Table \ref{table.adj.type}. We can see from the Table that all three types of the connection are important to ComQA, however, the inter-sentence connection is the most important one: removing $ A^{\text{inter}} $ will result in more than one point drop. On the other hand, the intra-sentence connection is less useful for the ComQA. We conjecture that the basic sentence representations have already been built from the BERT based sequential encoder, adding more intra-sentence interaction may not benefit a lot for the task. On the contrary, the inter-sentence connection and global connection introduce the specific inductive bias of the document structure to the model \cite{battaglia2018relational}, which helps achieve good performance on ComQA. The result also suggests that incorporating the sentence level information is very useful for document understanding.

	\subsection{Error Analysis}
	Finally, we can see from Table \ref{tab.result} there are still gaps between our model and human performance, to better understand the remaining challenges, we randomly sample 100 incorrectly predicted samples of the HGNN$ _{\text{base}} $ model from the test set, based on the error type, we classify them into 4 classes:
	
	1) \textbf{Redundancy}(41\%), we find the most common mistake our model has made is the redundancy. That is, our model is prone to select the uninformative sentences. Especially, our model sometimes selects the sentence at the beginning of the document which repeats the question. However, it is redundant from the perspective of QA.
	
	2) \textbf{Missed Sentence}(30\%), we find our model also prone to miss some important sentences. For the case where the document processing is not very well, i.e., the sentences are not perfectly segmented, some short but useful sentences are missed. 
	
	3) \textbf{Answerability}(24\%), another type of error is answerability. Although some extracted answers seem to be right, however, the answer itself is wrong, such as the counterfactual answers or too subjective answers. Resolving this type of error requires background knowledge, such as commonsense. 
	
	4) \textbf{Others}(5\%). The rest of the errors are due to various reasons such as ambiguous questions, incorrect labels, and so on.
	
	\section{Conclusion}
	In this paper, we study the compositional question answering where the answer is composed of discontiguous segments in the document. We present a large scale Chinese ComQA dataset containing more than 120k human-labeled questions. The data construction process has undergone rigid inspections to ensure high quality. To solve the ComQA problem, we propose a hierarchical graph neural networks that incorporate document graph structure to the model. We also devise two novel tasks, i.e., question selection and node selection, to pre-train the model. The proposed methods achieve significant improvement over previous methods. We also conduct several ablation studies to demonstrate the superiority of the proposed pre-training tasks and the graph structure. However, there is still a large gap between our model with human performance, suggesting that there is still room for improvement in ComQA.
      \section{Acknowledgments}
		We thank the anonymous reviewers for their insightful comments. We also appreciate the dedicated annotation efforts contributed by \begin{CJK}{UTF8}{gbsn}马琳、李潇如、张智程\end{CJK}, among others.
	\appendix
		\section{Appendix}
\begin{figure}
			\centering
			\includegraphics[width=1.0\linewidth]{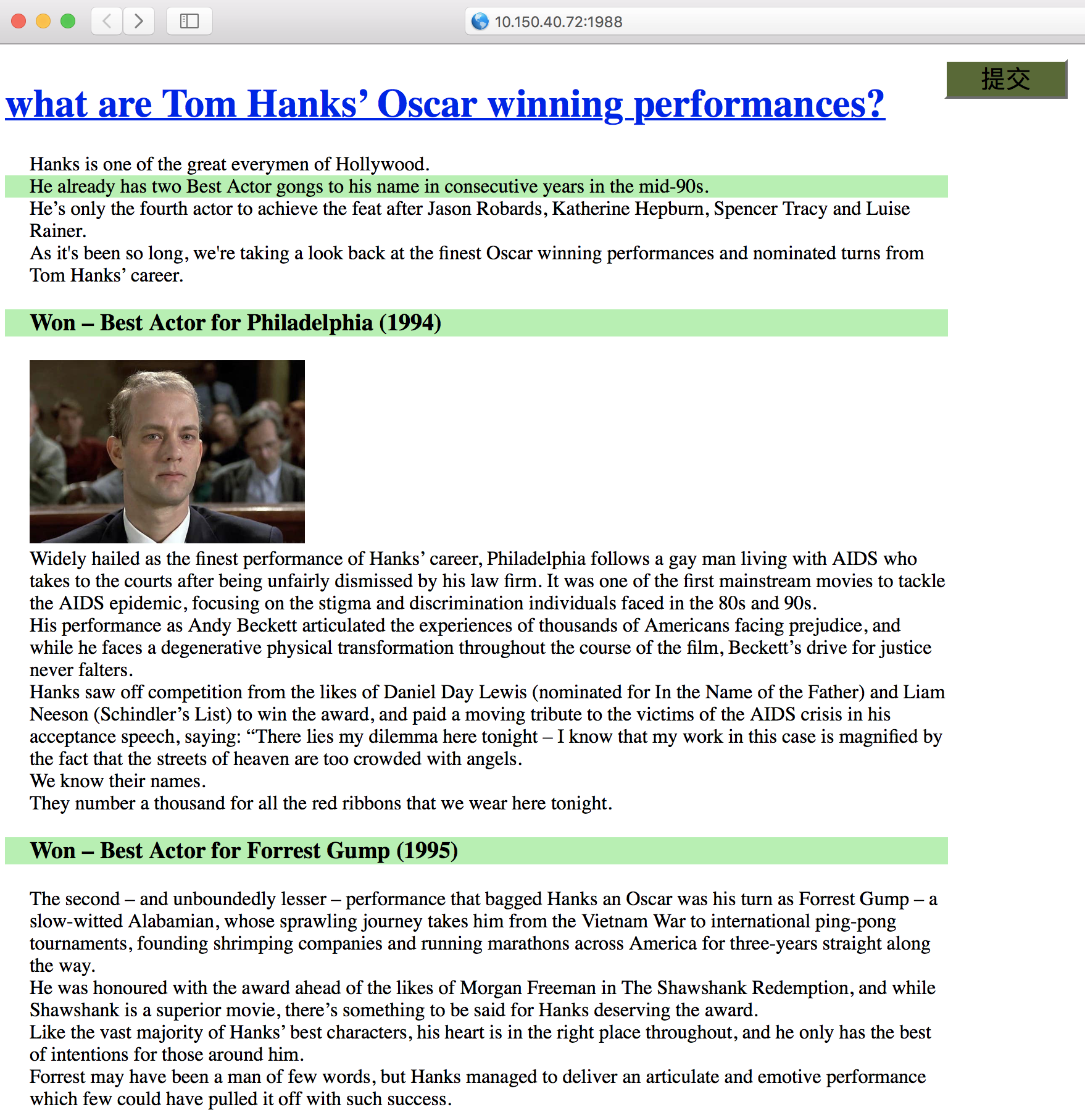}
			\caption{A snapshot of the annotation interface. Each row is a single node that could be served as the final answer component. The question is the page title in the top. Note that the image or table could also be selected.}\label{fig_label}
		\end{figure}

		The annotation interface is shown in Figure \ref{fig_label}. And the templates we used to determine weather a title of user query is a valid question is shown in Table \ref{table.template}.
		
		\begin{table}[!ht]
		\fbox{
			\small
			\centering
			\begin{minipage}{28em}
				\begin{CJK}{UTF8}{gbsn}为什么 (why),怎么回事 (what happened),什么情况 (what is the matter),啥情况 (what is the matter),咋回事 (what is wrong),原因 (reason),原理 (principle),由来 (origin),来由 (reason),会怎 (how will it be),为啥 (why),为何 (why),怎么还 (why still),怎么不 (why not),为甚 (why),为什 (why),看法 (opinion),评价 (evaluation),推荐 (recommendation),分享 (share),评测 (evaluate),排行 (rankings),排名 (ranking),对比 (compare),对待 (treat),区别 (distinguish),差异 (divergence),差别 (difference),不同 (different),好吗 (okay ？),好不好 (it is okay ？),有用吗 (is it useful ？),哪个好 (which is bettter ？),哪家好 (which one is better ？),比较好 (better),谁好 (who is good ？),谁厉害 (who is better ？),价值 (value),意义 (meaning),作用 (effect),用处 (use),功效 (function),危害 (harm),禁忌 (taboo),好处 (advantage),坏处 (disadvantage),优点 (advantage),缺点 (disadvantage),特点 (feature),特征 (characteristic),影响 (influence),哪些 (which),怎么样的 (how about it ？),哪 (where),还是 (still),怎么 (how),如何 (how about),怎样 (how),方法 (method),步骤 (step),操作 (operation),方案 (plan),办法 (method),教程 (course),方式 (way),玩法 (way of playing),攻略 (strategy),设置 (set up),自制 (homemade),做法 (action),过程 (process),流程 (process),规划 (planning),技巧 (skill),手续 (procedure),办理 (handle),规定 (regulation),要求 (demand),事项 (iitem),范围 (range),用什么 (by what),是谁 (who),谁是 (who is),谁最 (who is the best),什么 (what),多长 (how long),多少 (how many),多大 (how old),多重 (how heavy),多久 (how long),多远 (how far),多小 (how small),多块 (multiple blocks),多高 (how high),条件 (condition),介绍 (introduction),简介 (introduction),概况 (overview),简要 (brief),简明 (concise),意思 (meanning),标准 (standard),指标 (index),现状是 (the present situation is),什么叫 (what is called),何为 (what is),何谓 (what is),解释 (explanation),含义 (meaning),是否 (whether),能否 (can),可否 (can),是不是 (yes or no ?),会不会 (is not it ？),能不能 (can or not ？),有没有 (have or not),可不可 (can or not),吗 (is it ？),么 (what),哪 (where),几 (how many),有多 (how much)\end{CJK}	
			\end{minipage}
		}
		\caption{The templates we used to determine whether a title is a question when constructing the ComQA. There are 122 templates in total and the derived questions have more than 90\% accuracy.}\label{table.template}
	\end{table}

	\bibliographystyle{ACM-Reference-Format}
	\bibliography{www2021}
	
\end{document}